# A Multimodal Soft Gripper with Variable Stiffness and Variable Gripping Range Based on MASH Actuator

Dannuo Li[†], Xuanyi Zhou[†], Quan Xiong, and Chen-Hua Yeow, *Member, IEEE*

*Abstract*— Soft pneumatic actuators with integrated strain-limiting layers have emerged as predominant components in the field of soft gripper technology for several decades. However, owing to their intrinsic strain-limiting layer design, these soft grippers possess a singular gripping functionality, rendering them incapable of adapting to diverse gripping tasks with different strategies. Based on our previous work, we introduce a novel soft gripper that offers variable stiffness, an adjustable gripping range, and multifunctionality. The MASH actuator-based soft gripper can expand its gripping range up to threefold compared to the original configuration and ensures secure grip by enhancing stiffness when handling heavy objects. Moreover, it supports multitasking gripping through specific gripping strategy control.

## I. INTRODUCTION

Soft robots hold substantial promise for diverse applications, owing to their adaptability, exceptional compliance, and controllability[1]–[5]. These applications encompass gripping[6]–[10], locomotion[11]–[13], and human-robot interaction[9], [14]–[16]. As a dominant soft actuator in recent years, soft pneumatic actuators (SPA) have garnered significant research attention[17]–[20]. These pneumatic actuators find utility in functions like emulating artificial muscles, enabling soft exoskeletons, and enhancing soft grippers.

Soft pneumatic actuators primarily employ an integrated strain-limiting layer to constrain motion in specific directions, facilitating movements such as bending or twisting driven by pneumatic pressure[21], [22]. This soft pneumatic actuator with an integrated strain limiting layer is inexpensive to produce, and different motion modes of the actuator are realized by using strain limiting layers with different structural designs. However, SPA equipped with an integrated strain-limiting layer typically exhibit a singular motion pattern. Consequently, many soft grippers based on such soft pneumatic actuators often lack the robust gripping capability required for objects of varying sizes[23], [24]. Furthermore, due to the typically low rigidity of the materials used in these strain-limiting layers, these soft grippers struggle to handle heavy objects.

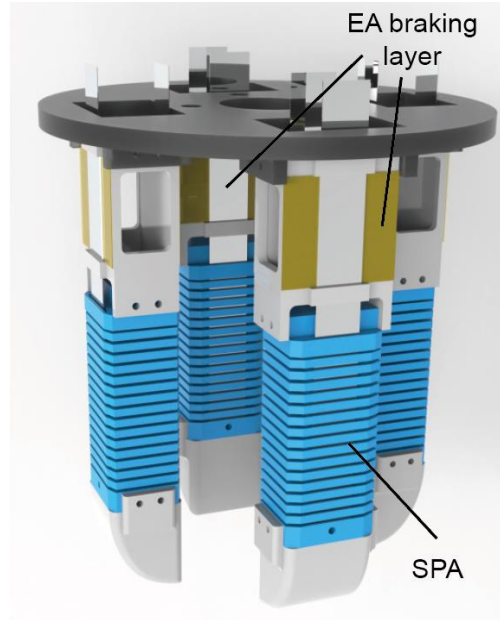

Fig. 1. Multimodal MASH-based soft gripper.

In our previous work, we proposed a Multi-Material, Adaptive Strain-Limiting, Hybrid Actuators (MASH)[25], [26]. This actuator can perform multiple motions by electrostatic adhesion of variable strain-limiting layer (EA Brake) action. When activating different EA braking layers, different braking forces in different directions will allow the SPA to bend or extend in different directions under pneumatic pressure.

Here, we propose a soft gripper based on MASH actuators and experimentally demonstrate its ability to achieve variable stiffness, variable range, and multitasking gripping. It consists of four MASH actuators and a 3D printed circular base, fingertip-like gripping component is installed at the end of each actuator(see Fig. 1). The MASH actuators are divided into two pairs and the internal chambers of the paired actuators are connected via air tubes. Those two pairs of actuators move independently of each other and can work together to realize multiple gripping modes under different control strategies.

This research is supported by National Robotics Programme–Robotics Enabling Capabilities and Technologies (W2025d0243). (Corresponding author: Chen-Hua Yeow)

[†] These authors contributed equally to this work.

Dannuo Li is with the Department of Biomedical Engineering and Advanced Robotics Centre, National University of Singapore, Singapore, 117583 Singapore (e-mail: ldannuo@gmail.com).

Xuanyi Zhou is with the Department of Biomedical Engineering, National University of Singapore, Singapore, 117583 Singapore (e-mail: e0978497@u.nus.edu).

Quan Xiong is with the Department of Biomedical Engineering and Advanced Robotics Centre, National University of Singapore, Singapore, 117583 Singapore (e-mail: e0788090@ u.nus.edu).

Chen-Hua Yeow is with the Department of Biomedical Engineering and Advanced Robotics Centre, National University of Singapore, Singapore, 117583 Singapore (e-mail: rayeow@nus.edu.sg).

Chen-Hua Yeow is also with Computer Science & Artificial Intelligence Laboratory, Massachusetts Institute of Technology, 02139, United States.

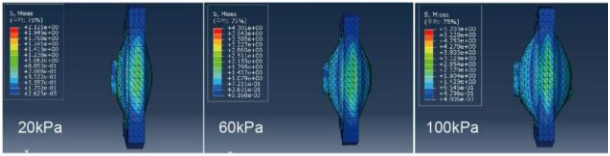

Fig. 2. Soft pneumatic actuator FEA simulation.

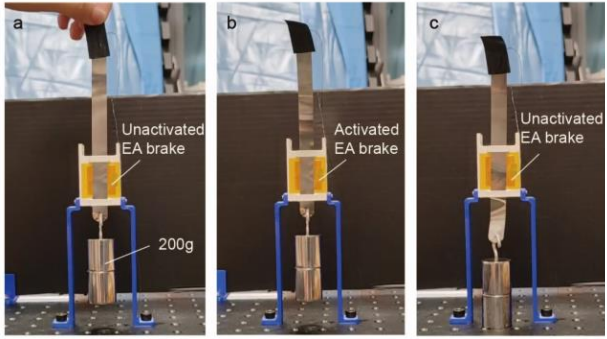

Fig. 3. The single side EA brake is capable of lifting loads of 200g at 20HZ, 2000V AC activation voltage. (a) Lifting the 200g load with hand. (b) Activating the EA brake. (c) Deactivating the EA brake.

## II. PRINCIPLE

### A. Electrostatic Adhesion Brake Mechanism

The EA brake acts as a strain limiting layer in the MASH actuators. Upon applying a voltage, it exerts a large restraining force on the soft pneumatic actuator. Upon application of the activation voltage, the copper piece that acts as the positive electrode in the EA brake and the externally grounded iron piece are adhere together by electrostatic forces. This adhesion force induces frictional coupling between the iron piece and the PI film within the EA brake, generating braking force.

Based on the parallel plate capacitor model, the maximum braking force that the EA brake can provide can be simply calculated by[27]:

$$F_{braking} = \mu \cdot \frac{\varepsilon_r \varepsilon_0 A U^2}{2d^2} \qquad (1)$$

Where $F_{braking}$ is the braking force, $\varepsilon_r$ is the relative permittivity(3.4) of the PI film acting as the dielectric layer. $\varepsilon_0$ is the permittivity of vacuum. A is the electrode contact area (3.175 $cm^2$) and μ (0.2) is the friction coefficient of the PI film surface.

### B. SPA Simulation

3D finite element analysis (FEA) models were made for the SPA internal unit chamber and analysis with ABAQUS/CAE (Simulia, Dassault Systems, RI). In the 3D FEA model, the material property was set to Homogeneous Ninjaflex and the coefficients were set to 2.61, -0.561 and 0.0972[28]. The meshing was done automatically using the tetrahedral hybridization formula of Abaqus with a grid size of 0.1 of the global size and a total number of 11,388 grids. The obtained relationship between pneumatic pressure and strain was fitted in MATLAB for comparison with experimental data.

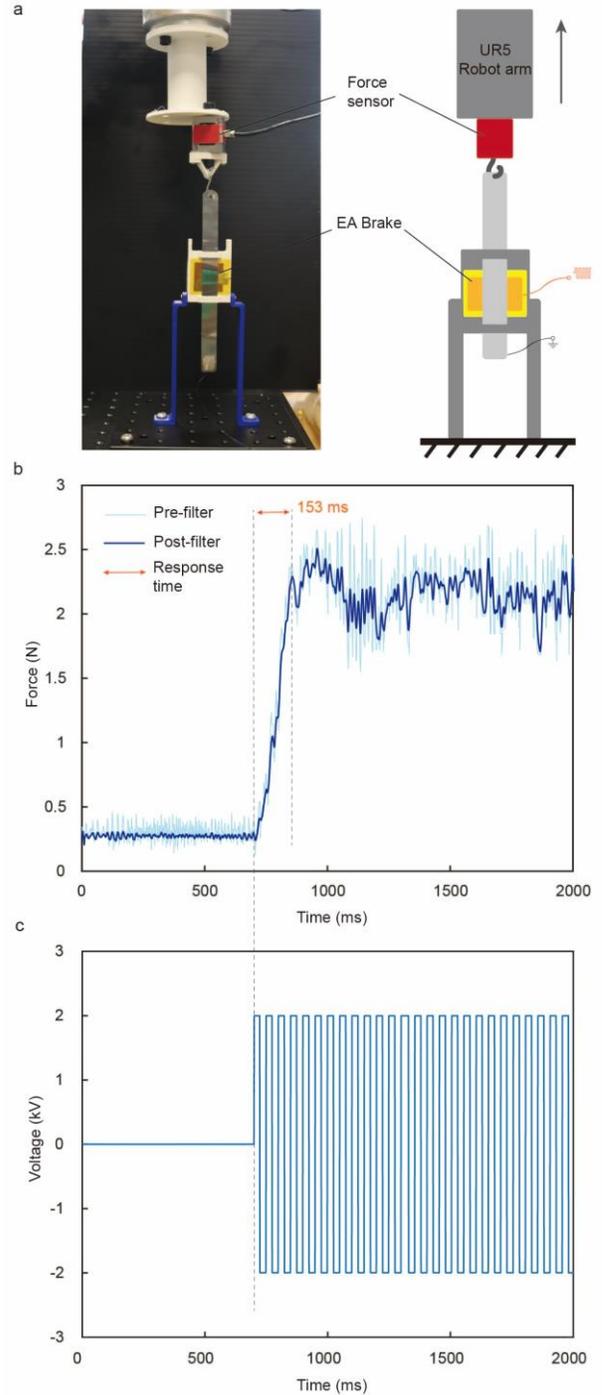

Fig. 4. EA brake activation response time experiment. (a) Experimental platform. (b) Restraining force changes before and after EA brake activation. (c) Activation voltage.

The FEA model shows that when pneumatic pressure is applied, the internal unit chambers of the SPA expand (see Fig. 2), causing them to unfold and result in the actuator's overall lengthening.

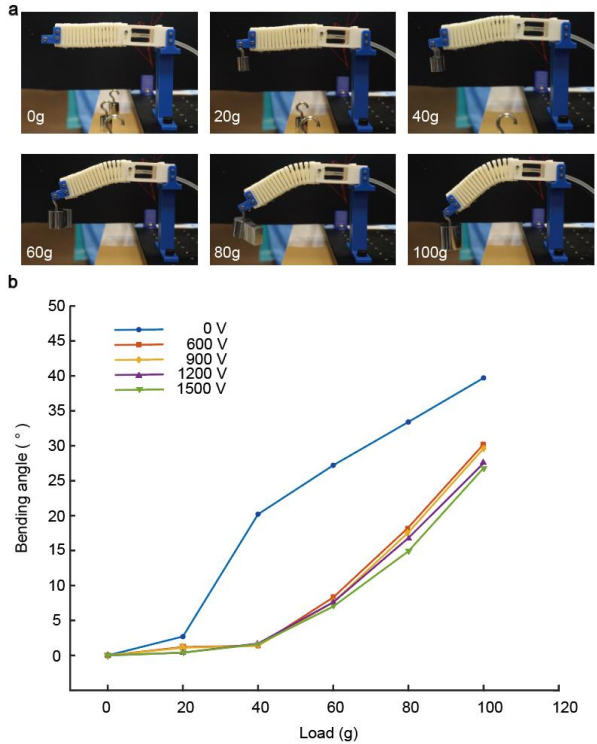

Fig. 5. Variable stiffness demonstration experiment. (a) Stiffness variation of SPA. (b) The measured variable stiffness.

## III. Design and Methods

### A. Multimodal MASH-based Soft Gripper Phototype

The body of the Multimodal MASH-based soft gripper consists of four MASH actuators with a circular base, where the detailed configuration parameters of the MASH actuators have been described in previous work. The four MASH actuators are mounted on the circular base in an equidistant circular array. These four MASH actuators are divided into two groups. The internal chambers of the two actuators in each group are connected via flexible air tubes, allowing for control at the same pneumatic pressure using the same regulator.

To enhance the gripping ability, we installed a fingertip-like gripping component at the terminus of each MASH actuator, fabricated by 3D printing soft material (TPU, Ninjaflex 95A) with an effective gripping area of 870 square centimeters.

### B. Multimodal Gripping Strategy

The Multimodal MASH-based soft gripper offers a range of gripping modes, and its gripping strategy can be adjusted for robust gripping depending on the different task.

*Small Single Object Gripping*

To begin, when an object enters the gripping range, the EA brake strain limiting layer is activated followed by suppling pneumatic pressure until the gripper firmly grips onto the object.

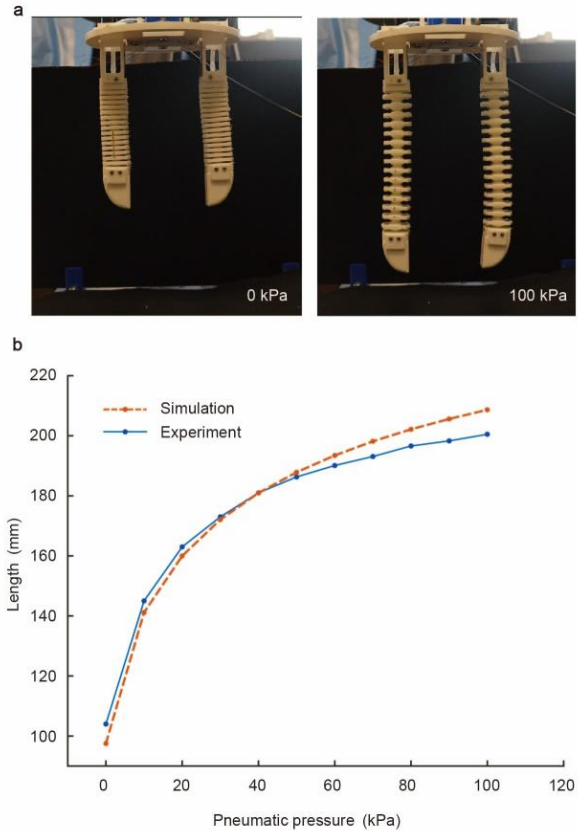

Fig. 6. Extended capability of SPA. (a) Morphology of SPA in the initial and end State. (b) Relationship between SPA length and input pneumatic pressure.

*Large Single Object Gripping*

Unlike smaller objects that is within the gripper grasp range when inactivated. Larger object requires to first increase the gripper grasp range by activating the outer EA brake strain-limiting layer of the MASH actuators followed by pneumatic pressure. These extents the grasp range. When the object fits the grasp range, the outer EA stain-limiting layer is released followed by activating the inner EA brake strain-limiting layer and pneumatic pressure to close the grasp onto the large object. This method of increasing the grasp range also enables the ability to grasp hollowed out object from their internal section as indicated in [25].

*Multi-object Gripping*

The soft pneumatic actuator structure in the MASH actuator gives the Multimodal MASH-based soft gripper the ability to multitasking gripping in three-dimensional vertical space by controlling 2 sets of actuators in tandem.

After the first set of MASH actuators has finished gripping the first object, pneumatic pressure is input to allow the second set of MASH actuators to extend in the vertical direction. After extension the second set of MASH actuators will be able to grip the second object.

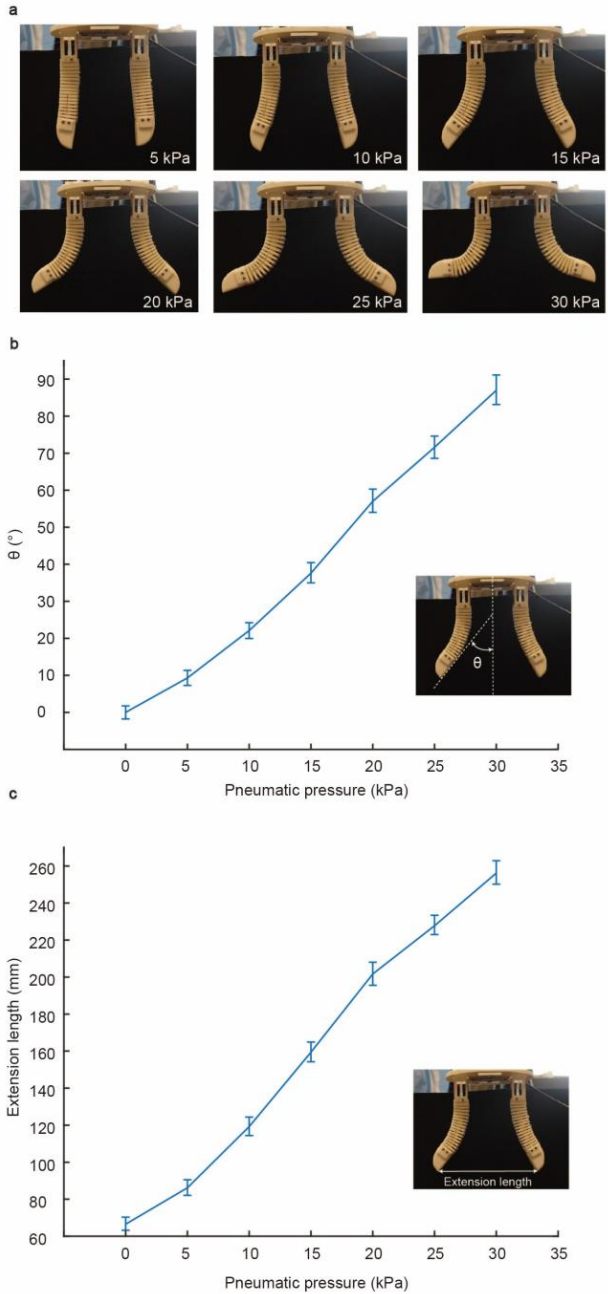

Fig. 7. Variable gripper range demonstration experiment. (a) Extension of SPA. (b) The measured extension angles. (c) The measured extension length.

## IV. EXPERIMENTAL AND CHARACTERIZATION

### A. Electrostatic Adhesion Brake Characterization

The power and maximum braking force of this EA brake under different conditions has been investigated in previous work[25], [26]. In this experiment the force output capacity of the EA brake was measured qualitatively at a frequency of 20 Hz AC with a voltage of 2000V. As shown in Fig. 3, the unilaterally activated EA brake was able to lift a 0.2kg weight with the applied voltage.

As shown in Fig. 4, the EA brake is activated with the steel piece pulled at a constant speed, and the variation curve of the restraining force is measured by the force sensor. The activation response time of the EA brake was measured to be in the range of 150~200ms after several experiments and obtaining a more ideal curve by limiting recursive averaging filtering.

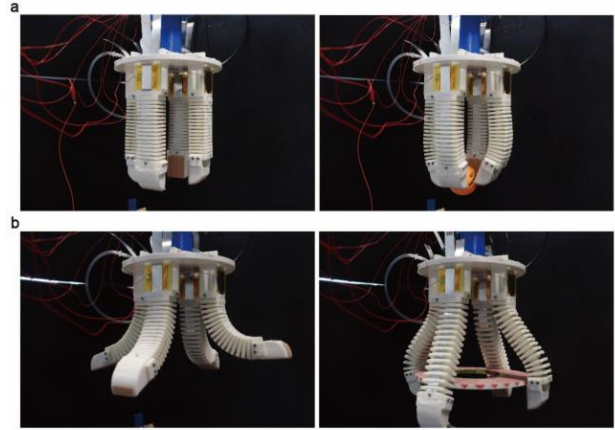

Fig. 8. Variable range gripping. (a) Small single object gripping. (b) Large single object gripping.

### B. Variable Stiffness

When activated by voltage applied to both sides of the EA brake, the soft pneumatic actuator in the MASH actuator changes in stiffness due to the influence of the EA brake's restraining force.

The variable stiffness of the EA brake was verified by applying different AC voltages in 300kV increments from 600kV with a frequency of 20Hz to the EA brake, and hanging various weights at 20g increments at the end of the MASH actuator as shown in Fig. 5.a. Measure the degree of deformation of the MASH actuator relative to the horizontal plane under different loads by using the image processing software(Tracker3.0), we obtain the results of the change in the stiffness of the soft pneumatic actuator in the MASH actuator under different activation voltages.

The results are shown in Fig. 5.b. With no activation voltage applied, the MASH actuator showed significant bending with the end-loaded weights. Whereas, applying activation voltage to the EA braking layer produced a significant stiffness change in the MASH actuator and the stiffness became larger as the activation voltage increased.

### C. Characterization of the SPA System

In conjunction with the EA brake and soft pneumatic actuator, the Multimodal MASH-based soft gripper enables variable range and dimensional gripping. The deformation range of the soft pneumatic actuator under different conditions is therefore an important performance metrics for Multimodal MASH-based soft gripper.

Since the configuration of each MASH actuator in the Multimodal MASH-based soft gripper is the same, to better demonstrate the deformation capability of the soft pneumatic actuator, in the following experiments we use the gripper experimental platform with only one pair of MASH actuators installed to conduct the experiment.

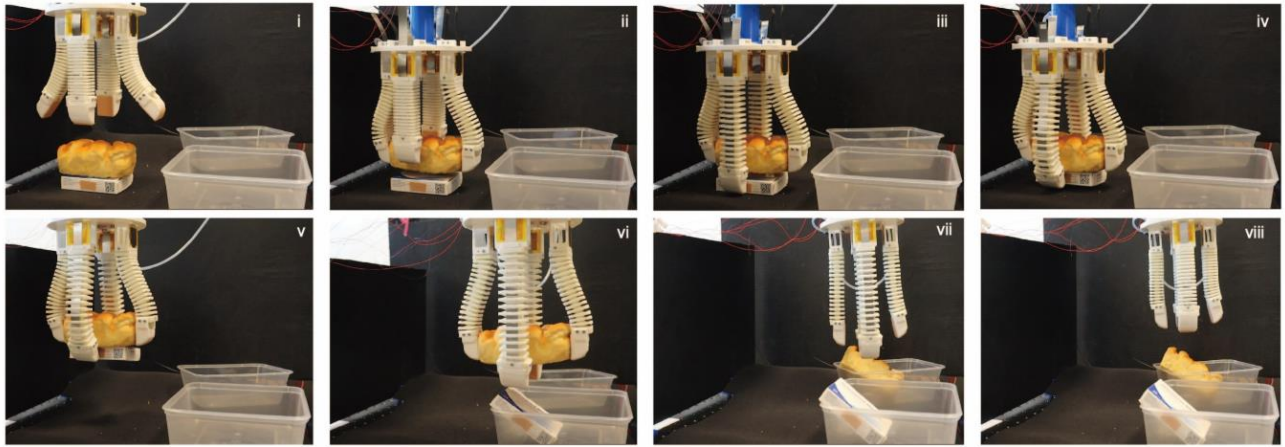

Fig. 9. Multitasking Gripping with Multimodal MASH-based soft gripper. (i) Activation the outer EA brakes of the first set of MASH actuators and applying pneumatic pressure. (ii) Activation the inner EA brake of the first set of MASH actuators, and then continue to input pneumatic pressure to complete the solid gripping. (iii) Applying pneumatic pressure to extend the second set of MASH actuators vertically. (iv) Activating the inner EA brakes of the second set of MASH actuators and applying pneumatic pressure. (v) Lifting of the two objects. (vi) Deactivation of the inner EA brake of the second set of MASH actuators. (vii) Deactivation of all of EA brake of the first set of MASH actuators. (viii) Stop input pneumatic pressure.

We explored the extension length of the soft pneumatic actuator at different pneumatic pressures without activating the EA brake. As shown in Fig. 6.a. In the initial state, the original length of the SPA in the MASH actuator was 104 mm. we entered several sets of pneumatic pressures (10-100 kPa) and measured the length of the SPA at different pressures. At 100kPa, the SPA can extend up to 200.55mm. As shown in Fig. 6.b, the experimentally obtained relationship of SPA elongation with pneumatic pressure input fits well with the relationship curve obtained from simulation.

The extent to which the soft pneumatic actuator can expand outwards when the EA brake applies an outer side restraint force determines the gripping range of the gripper in a single dimension. With a voltage applied to activate the outer limiting layer of the EA brake, we input pneumatic pressure to one pair of soft pneumatic actuators and observed the outward extension of the MASH actuators. We measured the angle of MASH actuator extension and the distance between the end of the actuators under six sets of gradually increasing pneumatic pressures (5-30kPa) using image processing software (Tracker 3.0). The extension distance between the ends of a pair of MASH actuators is 256mm at 30kPa pneumatic pressure, which is approximately three times the initial distance (see Fig. 7.a).

*D. Variable Range Gripping*

We demonstrate the ability of the Multimodal MASH-based soft gripper to grip objects at variable ranges. As shown in Fig. 8.a, activating all the EA braking layers on the inside of the gripper and inputting pneumatic pressure in the SPA, the four MASH actuators bend inward under the limiting force provided by the braking layers to complete the gripping of a table tennis ball with a radius of 2 cm.

For objects beyond the initial gripping range of the Multimodal MASH-based soft gripper, following the gripping strategy for large objects mentioned in the previous section, the soft gripper completes the gripping of a roll of tape with a radius of 7.5cm, which is 2.5 times the initial gripping radius (see Fig. 8.b).

*E. Multi-object Gripping*

With both pairs of MASH actuators acting, the Multimodal MASH-based soft gripper enables multi-object gripping. As shown in Fig. 9, after mounting the gripper on the UR5 robot arm, we performed a demonstration of the multi-object gripping process.

After activating the outer EA brakes of the first set of MASH actuators, pneumatic pressure is applied to expand this set of actuators outward, thus completing the subsequent grasping of the large Object 1. After the Object 1 enters the gripping range, activate the inner EA brake of this group of MASH actuators, and then continue to input pneumatic pressure to complete the solid gripping of the Object 1.

For the grasping of Object 2, it is first necessary to input pneumatic pressure to make the second set of MASH actuators extend in the vertical direction. After the Object 2 enters the gripping range, the inner EA brake of the second set of MASH actuators is activated and pneumatic pressure is continuously supplied until the object 2 is gripped.

## V. CONCLUSION

In this work, we present a variable stiffness variable range multimodal gripper based on MASH actuators. We explored the feasibility of applying the MASH actuator to the grasping task. We verify the variable stiffness due to the braking force provided by the EA brake layer. We also experimentally verified that this gripper can realize the variation of gripping range under the action of the EA braking layer in cooperation with the SPA. Finally, we verified its robust grasping capability by performing single small object, single large object, and multitasking gripping experiments using this gripper.

In the future, we can optimize the structural design of the EA braking layer in the MASH actuator so that it can provide greater braking force and variable stiffness at lower activation voltages. Explore the maximum input pneumatic pressure of SPA at different activation voltages and its variable stiffness limit capability. In the next phase, we can improve the gripping structure at the end of the actuator for better gripping performance.